\documentclass[10pt,conference]{IEEEtran}
\IEEEoverridecommandlockouts
\usepackage{cite}
\usepackage{multirow}
\usepackage{courier}
\usepackage{longtable}
\usepackage{subcaption}
\usepackage{hyperref}
\usepackage{xcolor}
\usepackage{graphicx}
\usepackage{booktabs}
\usepackage{makecell}

\definecolor{complete}{rgb}{0.0,0.7,0.0}
\definecolor{fwcolor}{rgb}{0.0,0.0,1.0}
\definecolor{rmcolor}{rgb}{0.7,0.0,0.0}

\newcommand{\Name}{CORL}
\newcommand{\highlevel}{H}
\newcommand{\midlevel}{M}
\newcommand{\lowlevel}{L}
\def\BibTeX{{\rm B\kern-.05em{\sc i\kern-.025em b}\kern-.08em
    T\kern-.1667em\lower.7ex\hbox{E}\kern-.125emX}}
\begin{document}

\title{Static Neural Compiler Optimization via \\ Deep Reinforcement Learning}

\author{\IEEEauthorblockN{Rahim Mammadli}
\IEEEauthorblockA{\textit{Technische Universität Darmstadt}\\
\textit{Graduate School of Excellence} \\
\textit{Computational Engineering}\\
mammadli@cs.tu-darmstadt.de}
\and
\IEEEauthorblockN{Ali Jannesari}
\IEEEauthorblockA{\textit{Iowa State University}\\
\textit{Department of Computer Science}\\
jannesari@iastate.edu}
\and
\IEEEauthorblockN{Felix Wolf}
\IEEEauthorblockA{\textit{Technische Universität Darmstadt}\\
\textit{Department of Computer Science} \\
wolf@cs.tu-darmstadt.de}
}

\maketitle

\begin{abstract}
  The phase-ordering problem of modern compilers has received a lot of attention from the research community over the years, yet remains largely unsolved. Various optimization sequences exposed to the user are manually designed by compiler developers. In designing such a sequence developers have to choose the set of optimization passes, their parameters and ordering within a sequence. Resulting sequences usually fall short of achieving optimal runtime for a given source code and may sometimes even degrade the performance when compared to unoptimized version. In this paper, we employ a deep reinforcement learning approach to the phase-ordering problem. Provided with sub-sequences constituting LLVM’s O3 sequence, our agent learns to outperform the O3 sequence on the set of source codes used for training and achieves competitive performance on the validation set, gaining up to 1.32x speedup on previously-unseen programs. Notably, our approach differs from autotuning methods by not depending on one or more test runs of the program for making successful optimization decisions. It has no dependence on any dynamic feature, but only on the statically-attainable intermediate representation of the source code. We believe that the models trained using our approach can be integrated into modern compilers as neural optimization agents, at first to complement, and eventually replace the hand-crafted optimization sequences.
\end{abstract}


\section{Introduction}
\label{sec:intro}

Code optimization remains one of the hardest problems of software engineering. Application developers usually rely on a compiler's ability to generate efficient code and rarely extend its standard optimization routines by selecting individual passes. The diverse set of applications and compute platforms make it very hard for compiler developers to produce a robust and effective optimization strategy. Modern compilers allow users to specify an optimization level which triggers a corresponding sequence of passes that is applied to the code. These passes are initialized with some pre-defined parameter values and are executed in a pre-defined order, regardless of the code being optimized. Intuitively, this rigidness limits the effectiveness of the optimization routine. Indeed, a recent study~\cite{Gong2018} has shown that even the highest optimization level of different compilers leaves plenty of room for improvement.

In order to establish an even playing field with existing optimization sequences, we limit the scope of our approach by allowing as input only information which is statically available during compilation. This sets us apart from autotuning approaches where the data gathered from one or multiple runs of the program is used to supplement the optimization strategy. Our predictive model uses the intermediate representation (IR) of the source code to evaluate and rank different optimization decisions. By iteratively following the suggestions of our model, we are able to produce an optimization strategy tailored to a given IR. This is the main difference of our approach from pre-defined optimization strategies shipped as part of modern compilers.

More formally, we rephrase the phase-ordering problem as a reinforcement learning problem. The environment is represented by the operating system and the LLVM optimizer. The agent is a deep residual neural network, which interacts with the environment by means of ~\textit{actions}. The actions can be of various levels of abstraction, but ultimately they translate to passes run by LLVM's optimizer on the IR. We will discuss actions in greater detail in Section~\ref{sec:approach}. The state information that is used by the agent to make predictions is represented by the IR of the source code and the history of actions that produces the IR. In response to actions, the environment returns the new state and the reward. The new state is produced by the LLVM optimizer which runs the selected pass(-es) on the current IR and produces the new IR. The reward is calculated by benchmarking the new IR and comparing its runtime to that of the original IR (i.e., a reduction in runtime produces a positive reward while an increase produces a negative one). Through interchanging steps of exploration and exploitation we train an agent that learns to correctly value the various optimization strategies.

We believe that the agents produced by our approach could be integrated into existing compilers alongside other routines, such as O2 or O3. Our agent outperforms O3 in multiple scenarios, achieving up to 1.32x speedup on previously-unseen programs, and can therefore be beneficial in a toolkit of optimization strategies offered to an application developer. While the agent learns to achieve superior performance on the training set, it is, on average, inferior to O3 on the validation set. However, we believe that this is due to current limitations of encoding we use for the IRs and the relatively small size of our dataset. We are convinced that the optimization strategies of the future will resemble learned agents rather than manually designed sequences, and that reinforcement learning will likely be the framework used to produce these agents. This work intends to be one of the first steps in this direction.

The static phase-ordering problem considered in this work is challenging because of several factors. First, the limited amount of information available during compilation, such as the unknown input size already reduces the optimization potential of the agent. In order to partially offset this we include benchmarks with various problem sizes in our dataset. Next, the number of possible optimization sequences grows exponentially with the number of passes. The space grows even more if we consider possible parameterizations of distinct passes. To deal with the large optimization space we employ several strategies: (i) we make the agent pick only a single action at a time instead of predicting the whole sequence from scratch, (ii) we experiment with different levels of abstraction for our actions, from triggering a sequence of optimization passes down to selecting a parameter for a single pass. Moreover, the space of possible IRs of each source code can also be quite large, therefore to encode the IRs we use the embeddings by Ben-Nun et al.~\cite{NIPS2018}. Another challenge is that the efficacy of different optimizations might vary depending on the underlying hardware. In our approach we use only one out of two available distinct system configurations per agent to run all of the benchmarks. This means that the learned agents are fine-tuned for the given hardware. However, this is not necessarily a disadvantage, because it is possible to train an agent once per processing unit and ship it alongside a compiler optimizer. Moreover, it could also be possible to train a single versatile agent by supplementing state with the information about the underlying hardware.

We use a dataset of 109 single-source benchmarks from the LLVM test suite to train and evaluate our model. The model is trained using IRs of source codes from the training set and evaluated on the source codes in the validation set. Using passes from the existing O3 sequence of the LLVM optimizer we are able to train an agent which is on average 2.24x faster than the unoptimized version of a program in the training set, whereas O3 is 2.17x faster. The best-performing agent on the validation set achieves an average of 2.38x speedup over the unoptimized version of the code, while the O3 sequence achieves an average of 2.67x speedup.

Most of the prior work related to ours~\cite{kulkarni2012mitigating, Ashouri:2017} focuses on the autotuning problem, where a program has to be run one or more times before it is possible to choose the optimization sequence. The advantage of these approaches is that the dynamic information gathered during program runs provides an accurate characterization of the program. These methods are therefore usually quite successful in outperforming compilers' pre-defined optimization sequences. However, a big disadvantage of these approaches is that they require extra developer effort to run the program and gather the necessary information, which prevents them from being integrated into compilers as part of a standard compilation routine. The supervised learning methods applied to compiler optimization problem require a pre-existing labeled dataset that is then used to train a model. Producing such a dataset is not an easy task because the search space is usually very large and the value of different data points is unkonwn beforehand. Reinformcement learning, in contrast to supervised learning, allows the trained agent to explore the environment and continuosly choose the data points itself as it learns. 
The problem of developing methods competing with pre-defined optimization sequences using only static information has not gained much attention in the scientific literature in recent years. This is partially because the problem is very challenging. Nonetheless, we believe that this problem is at least of equal importance and to the best of our knowledge we are the first to apply deep\footnote{Deep reinforcement learning encompasses a subset of reinforcement learning methods where the learning part is performed by a deep neural network.} reinforcement learning to solve it.
%
This paper makes the following contributions:
\begin{itemize}
  \item A novel deep reinforcement learning approach to static code optimization. The approach does not rely on manual feature engineering by a human, but instead learns by observing the effects of the various optimizations on the IR and the rewards from the environment. The approach is therefore fully automatic and relies only on the initial supply of the source codes.
  \item A trained optimization agent that can be integrated into modern compilers alongside existing optimization sequences exposed through compiler flags such as -O2, -O3, etc. The agent can produce IRs that are up to 1.32x faster than the ones resulting from using the O3 optimization sequence.
  \item An efficient framework "Compiler Optimization via Reinforcement Learning" (\Name{}) allowing fast exploration and exploitation in batches. The dynamic load-balancing mechanism distributes the benchmarking workload across the number of available workers and facilitates efficient exploration. Using a large replay memory allows for fast off-policy training of the agent. The results of benchmarks are further stored in a local database to allow reproducibility as well as higher efficiency of subsequent runs.
\end{itemize}

This paper is structured in the following manner. We start by providing background information in Section~\ref{sec:background}, before introducing our approach and~\Name{} framework in Section~\ref{sec:approach}. We evaluate our approach in Section~\ref{sec:evaluation} and describe the related work in Section~\ref{sec:related}. Finally, we conclude our paper and sketch future work in Section~\ref{sec:conclusion}.

\section{Background}
\label{sec:background}

Modern compilers expose multiple optimization levels via their command line interface. For example, the current version of LLVM\footnote{\url{https://releases.llvm.org/10.0.0/tools/clang/docs/CommandGuide/clang.html\#code-generation-options}, Access date: 22.06.2020} offers a selection of seven such levels. These aim to strike a certain trade-off between the size of the produced binary and its performance. Each optimization level corresponds to a unique sequence of passes that are run on the source code. These passes are constructed using hard-coded values matching the selected optimization level. Having to maintain multiple manually-designed optimization sequences is one of the drawbacks of the current design. Another disadvantage is that while the optimization sequences are generally efficient, they are not optimal, and in certain cases can even increase the runtime when compared to an unoptimized version. For example, after applying the O3 optimization sequence to the~\texttt{evalloop.c} benchmark from the LLVM test suite we observed a more than three-fold slowdown. In comparison to hand-crafted sequences of passes our method is fully-automatic and can learn to achieve any sort of a trade-off given the correct reward function. Constructing such a reward function for the size of the binary or the runtime is trivial, as discussed in Section~\ref{sub:problem_definition}.

The majority of existing machine learning methods to compiler optimization tackle the problem of autotuning. This means that they depend on dynamic runtime information and require at least one run of the source program to make a prediction. Apart from this, some of these methods rely on static features extracted from the source code, such as tokens~\cite{Cummins2017a}, IR~\cite{NIPS2018}, etc. In contrast to these methods we abandon the dependence on the dynamic features and aim to achieve the best-possible performance by only using the static features extracted from the IR~\cite{NIPS2018}.

\section{Approach}
\label{sec:approach}

We first give a high level overview of our approach in Section~\ref{sub:overview}, before formally defining the problem in Section~\ref{sub:problem_definition}. Then, we discuss the three levels of abstraction for actions we consider and the corresponding action spaces in Section~\ref{sub:action_spaces}, before introducing the tools used to map actions to concrete LLVM passes in Section~\ref{sub:implementation}. We finish by going over the functionality offered by the~\Name{} framework in Section~\ref{sub:corl_framework}.

\subsection{Overview}
\label{sub:overview}
\begin{figure}[htbp]
   \begin{center}
   \includegraphics[width=0.9\linewidth]{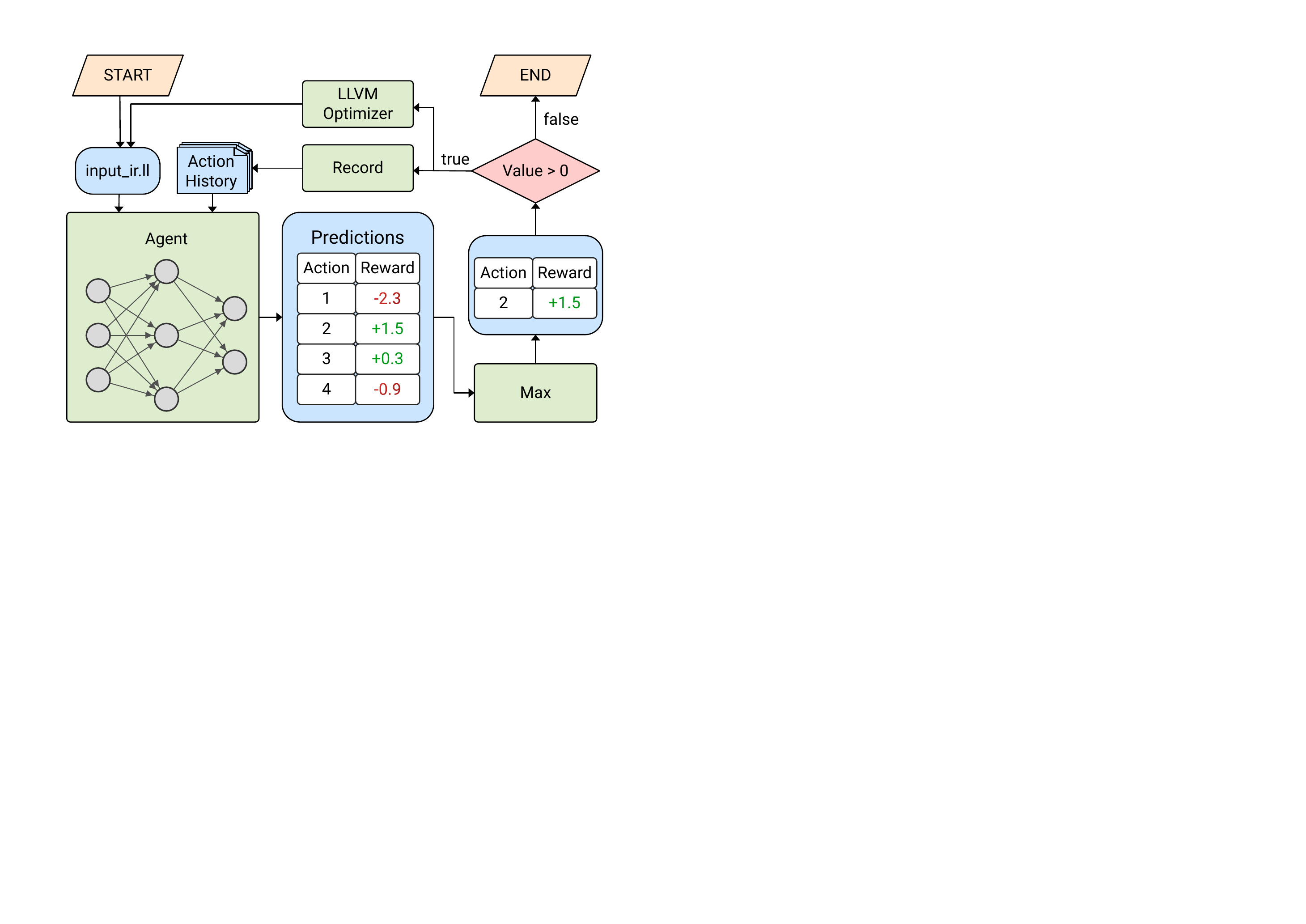}
   \end{center}
   \caption{The \Name{} workflow.}
   \label{fig:approach_overview}
\end{figure}

The high-level overview of our approach is shown in Figure~\ref{fig:approach_overview}. The agent takes the IR of the source code as well as the initially empty history of actions and calculates the expected cumulative rewards for different actions. If the predicted reward of an action is positive then it is expected to eventually lead to a speedup, while the negative rewards are expected to result in a slowdown. To ensure this, the rewards are calculated as $log(speedup)$ during training. Next, the action with the highest reward is chosen and if its value is positive LLVM optimizer applies the chosen optimization(s) to the input IR. The produced IR alongside the updated history of actions is then fed into the agent once again and the cycle continues. Eventually, the cycle breaks when the highest predicted reward is negative or the maximum number of allowed optimizations is applied. Having an upper-bound on the number of optimizations prevents the agent from potentially being stuck in an infinite loop.

\subsection{Problem Definition}
\label{sub:problem_definition}
We define the phase-ordering problem as a reinforcement learning problem. Since the IR of a source code carries only static information that we use to represent states, the states lack the Markovian property. Moreover, using embeddings produced by Ben-Nun et al.~\cite{NIPS2018} results in a further loss of information about the IR such as immediate values of instructions. To enrich the state representation we supplement it with the history of actions performed by the agent.

For a set of all possible states $S$ and actions $A$, our goal is to learn the value function $Q(s,a,w)$ parameterized with weights $w$, such that for any state $s \in S$ and action $a \in A$, the function predicts the highest cumulative reward attainable by taking that action. In order to learn the value function we enforce consistency:

\begin{equation}
Q(S_{t},A_{t},w) = R(S_{t},A_{t}) + \gamma \max_{a \in A} Q(S_{t+1},a,w)
\label{eq:bellmann}
\end{equation}

In the equation, $R(S_{t},A_{t})$ is the reward awarded for taking action $A_{t}$ in state $S_{t}$, after which the agent ends up in state $S_{t+1}$, and $\gamma \in [0, 1]$ is the discount factor for future rewards. Assuming that function $T(s)$ represents the runtime of the executable produced by compiling the IR corresponding to state $s \in S$, the reward for the action transitioning the agent from state $S_{t}$ to state $S_{t+1}$ is calculated as follows:

\begin{equation}
R = \ln{T(S_{t})\over T(S_{t+1})}
\end{equation}

Representing the reward as the logarithm of the attained speedup or slowdown allows for the rewards to be accumulated across transitions. Notably, to train an agent to minimize the size of the produced binary instead of its runtime, one would only need to update the reward function. Specifically, the function $T(s)$ calculating the runtime of an executable would need to be replaced with another function calculating its size. Similarly, using both the runtime and the size of an executable in the reward calculation would stimulate the agent to learn the trade-off between the two.

In order to learn an approximation of a function $Q(s,a,w)$ we first initialize a deep residual neural network (DQN) with random weights $w$. Then, we use a replay memory to sample experiences each represented as a set $\{S_t, A_t, R, S_{t+1}\}$ and compute the loss (TD-error) of our DQN as the squared mean of the difference between the left and right sides of Equation~\ref{eq:bellmann}.

\subsection{Action Spaces}
\label{sub:action_spaces}

\begin{figure}[htbp]
   \begin{center}
   \includegraphics[width=0.95\linewidth]{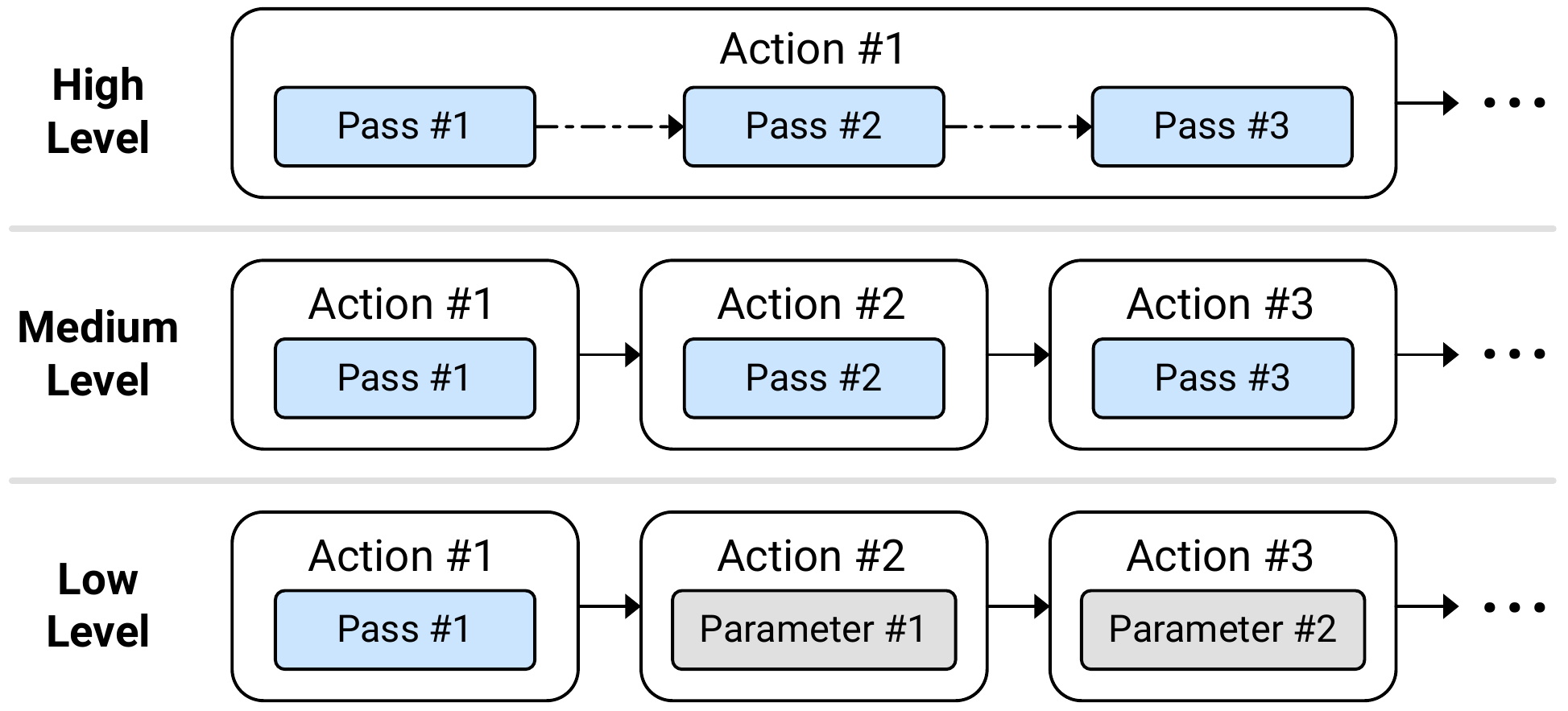}
   \end{center}
   \caption{Three levels of abstraction for actions.}
   \label{fig:action_spaces}
\end{figure}

In order to produce an optimization sequence for a given IR an agent must decide on a chain of actions. To represent the actions, we experiment with three different levels of abstraction, which are illustrated in Figure~\ref{fig:action_spaces}. At the highest level of abstraction an action triggers a series of passes to be applied to an IR. At the middle level of abstraction each action corresponds to an individual pass. Finally, at the lowest level of abstraction an action might select a pass or a parameter value for an already selected pass. For high and middle level actions the passes are initialized with pre-defined parameter values. The lower the level of abstraction for actions, the harder is the learning problem.

In this work we experiment with all three levels of abstraction. We label the action spaces produced by \textit{high}, \textit{middle} and \textit{low} level actions as \textit{\highlevel}, \textit{\midlevel}, \textit{\lowlevel} respectively. The size of each action space has exponential dependence on the maximum allowed number of consecutive actions, which we designate as parameter $\mu$. Selecting larger values for the parameters $\mu$ and $\gamma$ allows an agent to learn the existence of rewards lying many steps ahead. However, having $\mu$ too large may unnecessarily complicate the learning problem if such long-term dependences among actions do not exist. Furthermore, compilation time potentially also increases proportionally to $\mu$. Note that in action space~\lowlevel{} only actions selecting individual passes and not parameters contribute towards $\mu$. Moreover, since only the last parameter selection for every pass with multiple parameters makes it possible to construct and evaluate a pass, all the preceding intermediate actions produce a reward of 0. Therefore, to allow the agent to learn the values of different parameter selections of a given pass the value of $\gamma$ for all intermediate actions is set to 1.

\subsection{Implementation}
\label{sub:implementation}

Some of the passes in LLVM's O3 sequence are initialized with non-default constructors, and are therefore impossible to replicate using the command line interface of the LLVM optimizer $opt$. To allow experimentation at the highest level of abstraction in action space \highlevel{} using the exact passes from LLVM's O3 sequence, we create a special optimizer $opt\_corl$. This optimizer alters the functionality of $opt$ by using one or more user-specified subsequences of passes out of O3 to optimize a given IR. Each subsequence of passes is specified using its starting and ending indexes within the O3 sequence.

Both $opt\_corl$ and $opt$ allow experimentation in action space \midlevel{}. However, for the sake of generality, we only use $opt$ for the action spaces \midlevel{} and \lowlevel{}, since it allows us to specify both individual passes and set their parameters. When dealing with action space \lowlevel{}, $opt$ is only invoked when both pass and parameter selections have been finalized.

\subsection{\Name{} Framework}
\label{sub:corl_framework}


The majority of reinforcement learning algorithms can be described as iterative processes with interchanging exploration and exploitation steps performed in a loop. The sequential nature of these algorithms is usually not an issue for many reinforcement learning problems for which the exploration step completes in a short amount of time. Receiving a quick response to an action from the environment allows for fast generation of training data and consequently faster training~\cite{Mnih2013, silver2016mastering}. In contrast, for our problem the benchmarking step required to calculate the reward takes a relatively long time to complete. Therefore, waiting for the exploration step to finish before proceeding with the exploitation is suboptimal both in terms of the agent's training and efficient use of compute resources. To that end, we devise an algorithm which allows for the exploration and exploitation steps to be performed in parallel.

\begin{figure*}[htbp]
   \begin{center}
   \includegraphics[width=0.7\linewidth]{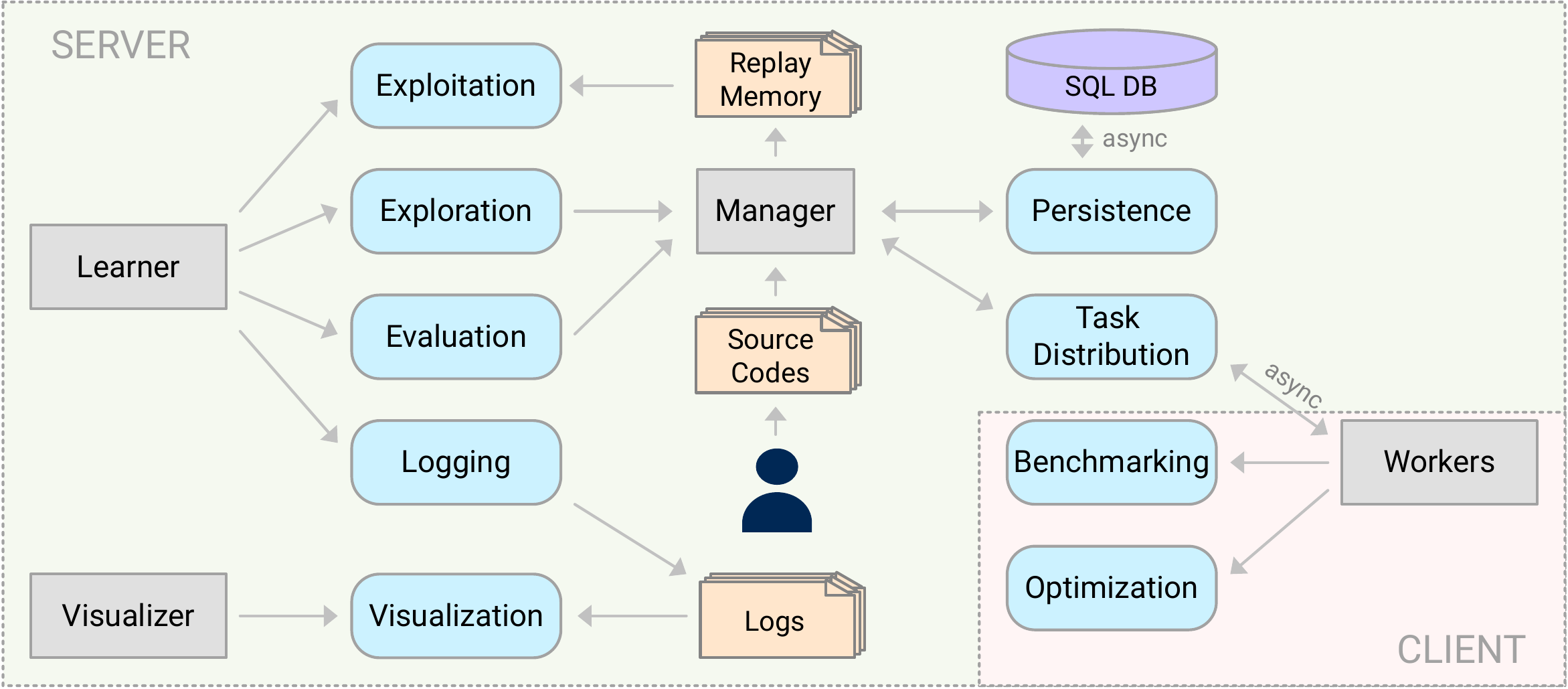}
   \end{center}
   \caption{Overview of the \Name{} framework.}
   \label{fig:corl_framework}
\end{figure*}

Figure~\ref{fig:corl_framework} illustrates the essential elements of the \Name{} framework, which is designed as a client-server architecture. The server-side functionality is divided across several objects responsible for training agents, managing workers and replay memory, and visualizing progress. As part of the exploration process the~\textit{learner} object generates new tasks in the form of state-action pairs and sends them to the~\textit{manager} object. Afterwards, as part of exploitation process, the learner continuously samples batches of experiences from the~\textit{replay memory} and trains the agent. The manager distributes the tasks generated by the learner across~\textit{workers} and updates the replay memory with the newly-generated experiences. Both the learner and the manager run in separate server-side processes, allowing for exploration and exploitation to be performed in parallel. Below we describe the various functionalities of the~\Name{} framework in a greater level of detail.


\subsubsection{Initialization}
\label{sub:corl_init}
The server-side logic starts with the manager scanning the source codes provided by the user and splitting them into training and validation sets. The programs are randomly shuffled and assigned to respective sets based on the user-specified ratio. Then, the manager loads previously-saved IRs and transitions from the SQL database into memory and populates the replay memory with experiences. Afterwards, workers are utilized to produce and benchmark unoptimized~\textit{base} IR and its O3-level optimized version for every source code in the dataset if not already present in memory. All the data generated at this stage and during exploration is asynchronously saved to the database. Upon completing this step the initialization is finished and the learner starts exploration.

\subsubsection{Benchmarking}
\label{sub:corl_benchmarking}
A single exploration step involves applying selected pass(-es) to a given IR, producing a new IR, which is then benchmarked to calculate the reward. Benchmarking any program is prone to noise and based on our observations the variation in runtime in terms of percentage of deviation from the mean is itself dependent on runtime. For the source codes in our dataset the bigger the runtime the smaller is the observed variation. Therefore, to calculate the runtime of an IR a worker runs it between 20 and 1000 times, depending on its runtime, and sends the median runtime back to the manager. To hide the latency induced by benchmarking, we perform exploration in batches. 

\subsubsection{Exploration}
\label{sub:corl_exploration}
To perform exploration we use an $\epsilon$-greedy strategy. The value of $\epsilon$ is linearly-annealed throughout training. The agent starts every exploration step by sampling a batch of~\textit{base} states. These states correspond to unoptimized versions of the IR for every source code in the training set. For each sampled state an agent selects an action either greedily or randomly based on the toss of a coin. State-action pairs already present in memory are used to perform server-side state transitions and the exploration proceeds with the new state until the transition for a selected action is not yet present in memory. Finally, an assembled set of state-action pairs is sent to the manager and the agent proceeds to exploitation.

\subsubsection{Exploitation}
\label{sub:corl_exploitation}
Before the exploitation process starts the replay memory has to be populated with a sufficient number of experiences. Once the replay memory is large enough, the learner starts to train the agent by minimizing the loss function described previously in Section \ref{sub:problem_definition}. To stabilize training we use fixed Q-targets which are updated once every $\tau$ steps. Every $\delta$ steps, where $\delta$ is a multiple of $\tau$, the framework switches to evaluation mode, during which both exploration and exploitation halts and the agent's performance is evaluated.

\subsubsection{Evaluation, Logging, and Visualization}
\label{sub:corl_evaluation}
Evaluation is performed similarly to exploration, with two main differences. First, instead of sampling base states from the training set, the agent is evaluated in all of the base states in the dataset, including the validation set. Second, instead of letting the toss of a coin determine the chosen action, the agent always behaves greedily. The learner logs all of the data pertaining to a single run of the CORL framework, including evaluation and exploitation progress, to a separate file. The \textit{visualizer} object continuously scans the logs directory and performs visualization using the VisDom framework\footnote{\url{https://github.com/facebookresearch/visdom}}.

\section{Evaluation}
\label{sec:evaluation}

We first explain the experimental setup in
Section~\ref{sub:experimental_setup}, before discussing the quality of
the fit achieved by our agents in Section~\ref{sub:convergence}. Then,
we introduce the metrics we use to evaluate the performance of our
agents in Section~\ref{sub:metrics}. We conclude with reviewing the results of our evaluation in Sections~\ref{sub:aggregate_results} and~\ref{sub:individual_performance}.

\subsection{Experimental Setup}
\label{sub:experimental_setup}
The dataset for training our optimizing agents consists of 109
single-source benchmarks from the LLVM test suite. Single-source
benchmarks were chosen as they provide a simple and convenient way of
building and executing the benchmarks. The complete list of benchmarks
and source codes is available in
Table~\ref{tab:list_of_benchmarks}. The programs were split between training
and validation sets in a 4:1 ratio. To speed-up experimentation we
excluded top-four source codes with the longest runtime when dealing
with action spaces~\midlevel{} and~\lowlevel{}.


{ 
\renewcommand{\arraystretch}{1}
\begin{table}
  \centering
  \caption{The list of benchmarks and source codes used for evaluation.}
  \label{tab:list_of_benchmarks}
  \footnotesize
  \begin{tabular}{l | p{21em}}
  \toprule
  Benchmark & Sources \\
  \midrule
  Polybench     & correlation.c covariance.c 2mm.c 3mm.c atax.c bicg.c cholesky.c doitgen.c gemm.c gemver.c gesummv.c mvt.c symm.c syr2k.c syrk.c trisolv.c trmm.c durbin.c dynprog.c gramschmidt.c lu.c ludcmp.c floyd-warshall.c reg\_detect.c adi.c fdtd-2d.c fdtd-apml.c jacobi-1d-imper.c jacobi-2d-imper.c seidel-2d.c \\
  \midrule
  Shootout      & ackermann.c ary3.c fib2.c hash.c heapsort.c lists.c matrix.c methcall.c nestedloop.c objinst.c random.c sieve.c strcat.c ackermann.cpp fibo.cpp heapsort.cpp matrix.cpp methcall.cpp random.cpp except.cpp \\
  \midrule
  Misc          & dt.c evalloop.c fbench.c ffbench.c flops-1.c flops-2.c flops-3.c flops-4.c flops-5.c flops-6.c flops-7.c flops-8.c flops.c fp-convert.c himenobmtxpa.c lowercase.c mandel-2.c mandel.c matmul\_f64\_4x4.c oourafft.c perlin.c pi.c ReedSolomon.c revertBits.c richards\_benchmark.c salsa20.c whetstone.c mandel-text.cpp oopack\_v1p8.cpp sphereflake.cpp \\
  \midrule
  Stanford      & Bubblesort.c FloatMM.c IntMM.c Oscar.c Perm.c Puzzle.c Queens.c Quicksort.c RealMM.c Towers.c Treesort.c \\
  \midrule
  BenchmarkGame & fannkuch.c n-body.c nsieve-bits.c partialsums.c puzzle.c recursive.c spectral-norm.c fasta.c \\
  \midrule
  Linpack       & linpack-pc.c \\
  \midrule
  McGill        & chomp.c misr.c queens.c \\
  \midrule
  Dhrystone     & dry.c fldry.c \\
  \midrule
  CoyoteBench   & almabench.c huffbench.c lpbench.c \\
  \midrule
  SmallPT       & smallpt.cpp \\
  \bottomrule

  \end{tabular}
\end{table}
}

To execute optimization sequences we use LLVM optimizer version 3.8. The choice of this particular version is motivated by the ability to use pre-trained embeddings from the study by Ben-Nun et al.~\cite{NIPS2018}. However, our approach can be used with the newer versions of the LLVM optimizer as well.

To define the action space~\highlevel{}, we partition the O3 sequence
of the LLVM optimizer into eight different actions, as shown in
Table~\ref{tab:o3_sequence_actions_h}. The division follows the
observation that the optimization sequence consists of smaller logical
sub-sequences ending with a \texttt{simplifycfg} pass. To allow a fair
comparison of the results of our experiments we define the actions in
spaces~\midlevel{} and~\lowlevel{} using 42 unique transformation
passes which are part of actions in space~\highlevel{}. In action
space~\midlevel{}, the passes are initialized with the default
parameter values, while in action space~\lowlevel{} agents also choose
the parameter values. Table~\ref{tab:tunable_parameters} lists the
passes in action space~\lowlevel{} which have tunable parameters along
with the values for these parameters. The value $\mu=16$, the maximum
number of actions, is used in all of the experiments.

{ 
\renewcommand{\arraystretch}{1}
\begin{table}
  \centering
  \caption{Passes within the O3 sequence of the LLVM optimizer version
    3.8, divided into eight different actions for experiments in the action space~\highlevel{}.}
  \label{tab:o3_sequence_actions_h}
  \footnotesize
  \setlength\tabcolsep{1.0pt}
  \begin{tabular}[t]{ l | l | l }
    \toprule
    Order & Pass & Action \\
    \midrule
    0	  & tti & \multirow{8}{*}{0} \\
    1	  & verify &  \\
    2	  & tbaa &  \\
    3	  & scoped-noalias &  \\
    4	  & simplifycfg &  \\
    5	  & sroa &  \\
    6	  & early-cse &  \\
    7	  & lower-expect &  \\
    \midrule
    8	  & targetlibinfo & \multirow{12}{*}{1} \\
    9	  & tti & \\
    10	& forceattrs & \\
    11	& tbaa & \\
    12	& scoped-noalias & \\
    13	& inferattrs & \\
    14	& ipsccp & \\
    15	& globalopt & \\
    16	& mem2reg & \\
    17	& deadargelim & \\
    18	& instcombine & \\
    19	& simplifycfg & \\
    \midrule
    20	& globals-aa & \multirow{10}{*}{2} \\
    21	& prune-eh & \\
    22	& inline & \\
    23	& functionattrs & \\
    24	& argpromotion & \\
    25	& sroa & \\
    26	& early-cse & \\
    27	& jump-threading & \\
    28	& correlated-propagation & \\
    29	& simplifycfg & \\
    \midrule
    30	& instcombine &  \multirow{3}{*}{3} \\
    31	& tailcallelim & \\
    32	& simplifycfg & \\
    \midrule
    33	& reassociate & \multirow{5}{*}{4} \\
    34	& loop-rotate & \\
    35	& licm & \\
    36	& loop-unswitch & \\
    37	& simplifycfg & \\
    \bottomrule

  \end{tabular}
  \begin{tabular}[t]{ l | l | l }
    \toprule
    Order & Pass & Action \\
    \midrule
    38	& instcombine & \multirow{17}{*}{5} \\
    39	& indvars & \\
    40	& loop-idiom & \\
    41	& loop-deletion & \\
    42	& loop-unroll & \\
    43	& mldst-motion & \\
    44	& gvn & \\
    45	& memcpyopt & \\
    46	& sccp & \\
    47	& bdce & \\
    48	& instcombine & \\
    49	& jump-threading & \\
    50	& correlated-propagation & \\
    51	& dse &  \\
    52	& licm & \\
    53	& adce &  \\
    54	& simplifycfg & \\
    \midrule
    55	& instcombine &  \multirow{11}{*}{6} \\
    56	& barrier & \\
    57	& rpo-functionattrs & \\
    58	& elim-avail-extern & \\
    59	& globals-aa & \\
    60	& float2int &  \\
    61	& loop-rotate & \\
    62	& loop-vectorize & \\
    63	& instcombine &  \\
    64	& slp-vectorizer & \\
    65	& simplifycfg &  \\
    \midrule
    66	& instcombine &  \multirow{8}{*}{7} \\
    67	& loop-unroll & \\
    68	& instcombine &  \\
    69	& licm & \\
    70	& \makecell[l]{alignment-from\\-assumptions} & \\
    71	& strip-dead-prototypes & \\
    72	& globaldce & \\
    73	& constmerge & \\
    \midrule
    \rule{0pt}{11.3pt}	&  &  \\
    \bottomrule
  \end{tabular}
\end{table}
}


{ 
\begin{table*}
  \centering
  \caption{Passes in the action space~\lowlevel{} that have tunable
    parameters. The first value is the default for each parameter.}
  \label{tab:tunable_parameters}
  \footnotesize
  \setlength\tabcolsep{1.2pt}
  \begin{tabular}[t]{l  l  l }
    \toprule
    Pass & Parameter & Values \\
    \toprule
    \rule{0pt}{11pt}\multirow{13}{*}{loop-vectorize} & vectorizer-maximize-bandwidth & [false, true] \\
     & max-interleave-group-factor & [8, 6, 10] \\
     & enable-interleaved-mem-accesses & [false, true] \\
     & vectorizer-min-trip-count & [16, 8, 32, 64] \\
     & enable-mem-access-versioning & [true, false] \\
     & max-nested-scalar-reduction-interleave & [2, 1, 4] \\
     & enable-cond-stores-vec & [false, true] \\
     & enable-ind-var-reg-heur & [true, false] \\
     & vectorize-num-stores-pred & [1, 2, 4] \\
     & enable-if-conversion & [true, false] \\
     & enable-loadstore-runtime-interleave & [true, false] \\
     & loop-vectorize-with-block-frequency & [false, true] \\
     & small-loop-cost & [20, 10, 30] \\
     \midrule
    \rule{0pt}{11pt}\multirow{9}{*}{simplifycfg}	& bonus-inst-threshold & [1, 2] \\
    	& phi-node-folding-threshold & [2, 3, 4] \\
    	& simplifycfg-dup-ret & [false, true] \\
    	& simplifycfg-sink-common & [true, false] \\
    	& simplifycfg-hoist-cond-stores & [true, false] \\
    	& simplifycfg-merge-cond-stores & [true, false] \\
    	& simplifycfg-merge-cond-stores-aggressively & [false, true] \\
    	& speculate-one-expensive-inst & [true, false] \\
    	& max-speculation-depth & [10, 5, 20] \\
      \midrule
      \rule{0pt}{11.7pt}\multirow{6}{*}{loop-unroll} & percent-dynamic-cost-saved-threshold & [20, 15, 25] \\
       & runtime & [false, true] \\
       & allow-partial & [false, true] \\
       & max-iteration-count-to-analyze & [0, 10, 100, 1000, 10000] \\
       & dynamic-cost-savings-discount & [2000, 1500, 2500] \\
       & threshold & [150, 75, 300, 600] \\

    \bottomrule
  \end{tabular}
  \begin{tabular}[t]{l  l  l }
    \toprule
    Pass & Parameter & Values \\
    \toprule
    \multirow{5}{*}{slp-vectorizer} & slp-vectorize-hor & [true, false] \\
     & slp-threshold & [0, 1, 2] \\
     & slp-vectorize-hor-store & [false, true] \\
     & slp-max-reg-size & [128, 64, 256, 512] \\
     & slp-schedule-budget & [100000, 50000, 200000] \\
    \midrule
 \multirow{3}{*}{inline} & inlinecold-threshold & [275, 175, 225, 325, 400] \\
  & inline-threshold & [275, 175, 225, 325, 400] \\
  & inlinehint-threshold & [325, 175, 275, 225, 400] \\
  \midrule
 \multirow{3}{*}{loop-unswitch} & with-block-frequency & [false, true] \\
  & threshold & [100, 60, 140] \\
  & coldness-threshold & [1, 2, 3] \\
  \midrule
 \multirow{3}{*}{indvars} & liv-reduce & [true, false] \\
  & verify-indvars & [true, false] \\
  & replexitval & [cheap, never, always] \\
  \midrule
 \multirow{3}{*}{gvn} & enable-pre & [true, false] \\
  & enable-load-pre & [true, false] \\
  & max-recurse-depth & [1000, 2000, 3000] \\
  \midrule
 \multirow{2}{*}{sroa} & sroa-random-shuffle-slices & [false, true] \\
  & sroa-strict-inbounds & [false, true] \\
  \midrule
  \multirow{2}{*}{jump-threading} & implication-search-threshold & [3, 2, 4] \\
   & threshold & [6, 3, 9, 12] \\
   \midrule
  loop-rotate & rotation-max-header-size & [16, 8, 32, 64] \\
  \midrule
  licm & disable-licm-promotion & [true, false] \\
  \midrule
 lower-expect & likely-branch-weight & [64, 32, 128] \\
 \midrule
 float2int & float2int-max-integer-bw & [64, 32, 128] \\

 \bottomrule
  \end{tabular}
\end{table*}
}

For our experiments, we run the server-side and the client-side logic of the CORL framework on two different hardware architectures. Below we describe these architectures in detail.
\subsubsection{Server}
The server-side logic responsible for training the agents and distributing tasks to clients was run on a single server with two Intel(R) Xeon(R) Gold 6126 2.60GHz CPUs, 64GBs of main memory, two NVIDIA GeForce GTX 1080 Ti GPUs, and Ubuntu 16.04 LTS operating system. We trained our models using a single GPU.

\subsubsection{Client}
We ran the clients on the nodes of the Hardware Phases I and II of the Lichtenberg High Performance Computer. The nodes within Hardware Phases I and II each have two Intel(R) Xeon(R) E5-2670 CPUs and Intel(R) Xeon(R) E5-2680 v3 CPUs respectively, 64GBs of main memory, and run CentOS Linux version 7. Each node ran a single client at a time, with the number of clients dynamically changing throughout the runs as the workers were added and removed from the pool. Due to availability constraints experiments with action space~\highlevel{} were performed on the nodes of Hardware Phase II while experiments with action spaces~\midlevel{} and ~\lowlevel{} were performed on the nodes of Hardware Phase I.

\subsection{Convergence}
\label{sub:convergence}

\begin{figure*}%
\centering
\scalebox{.8}
{
\begin{subfigure}{.32\textwidth}
  \centering
  \caption{Action space \highlevel{}}
  \includegraphics[width=1\linewidth]{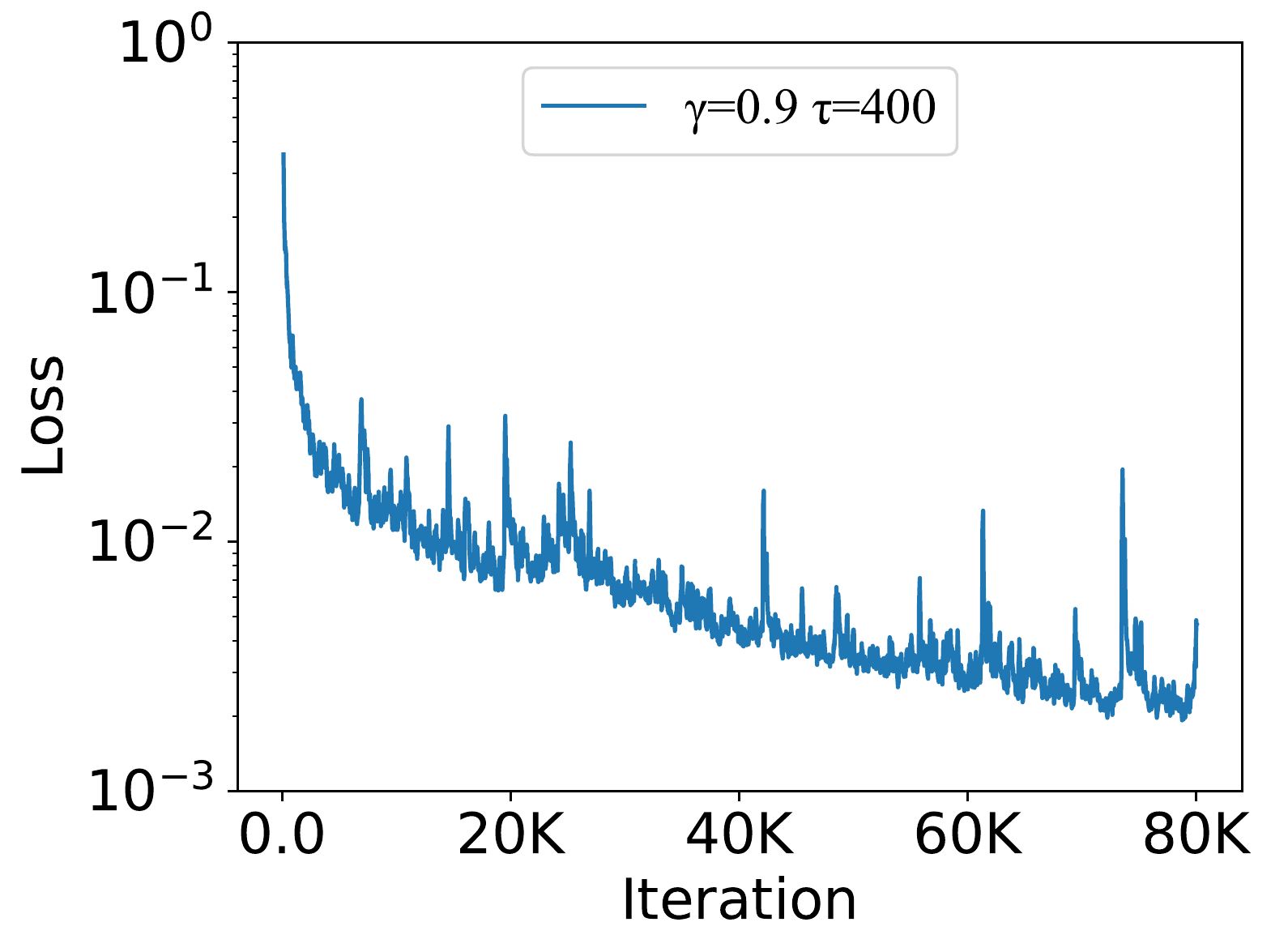}
  \label{fig:convergence_h}
\end{subfigure}\hspace{0.05\textwidth}
\begin{subfigure}{.32\textwidth}
  \centering
  \caption{Action space \midlevel{}}
  \includegraphics[width=1\linewidth]{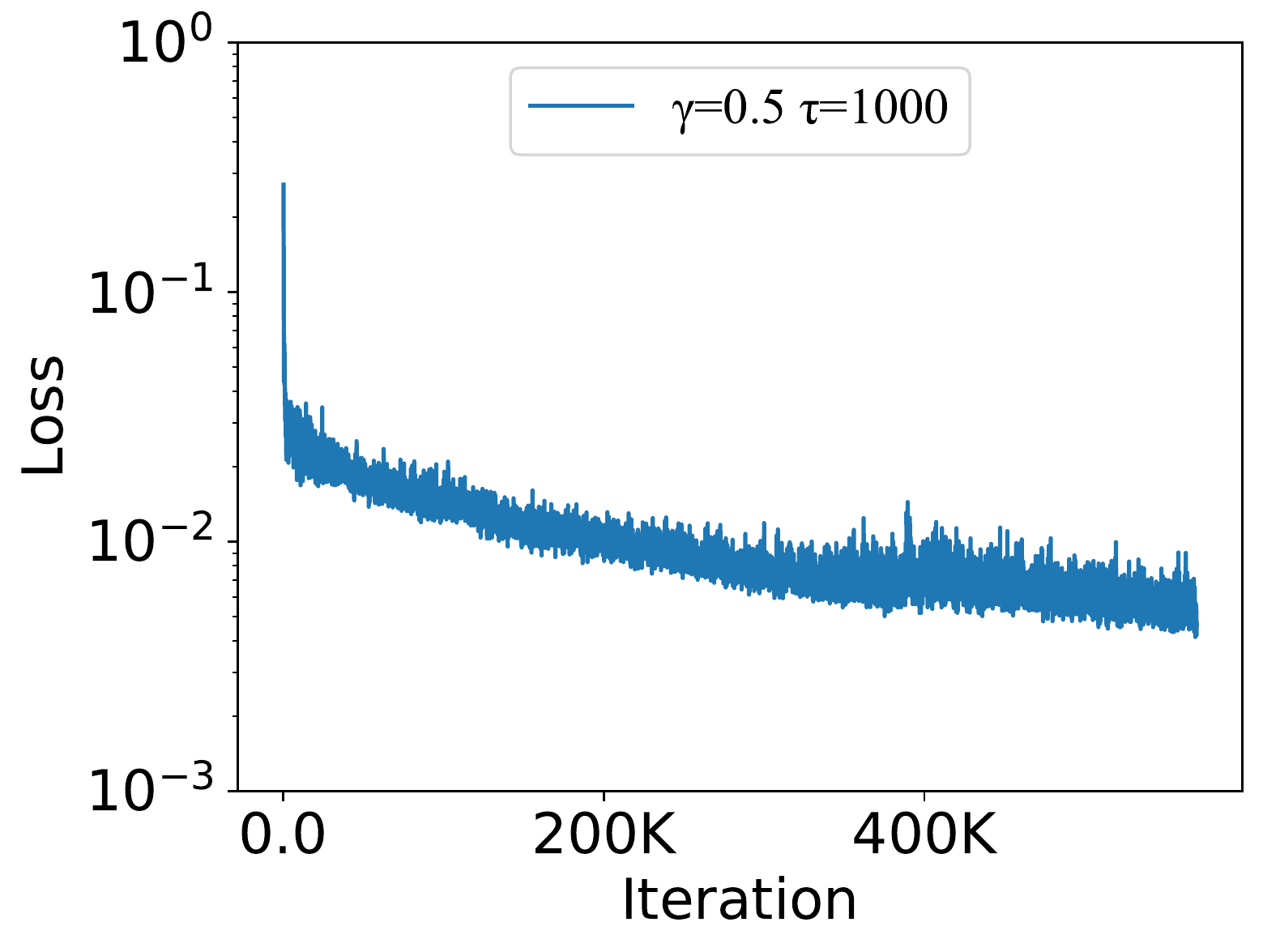}
  \label{fig:convergence_m}
\end{subfigure}\hspace{0.05\textwidth}
\begin{subfigure}{.32\textwidth}
  \centering
  \caption{Action space \lowlevel{}}
  \includegraphics[width=1\linewidth]{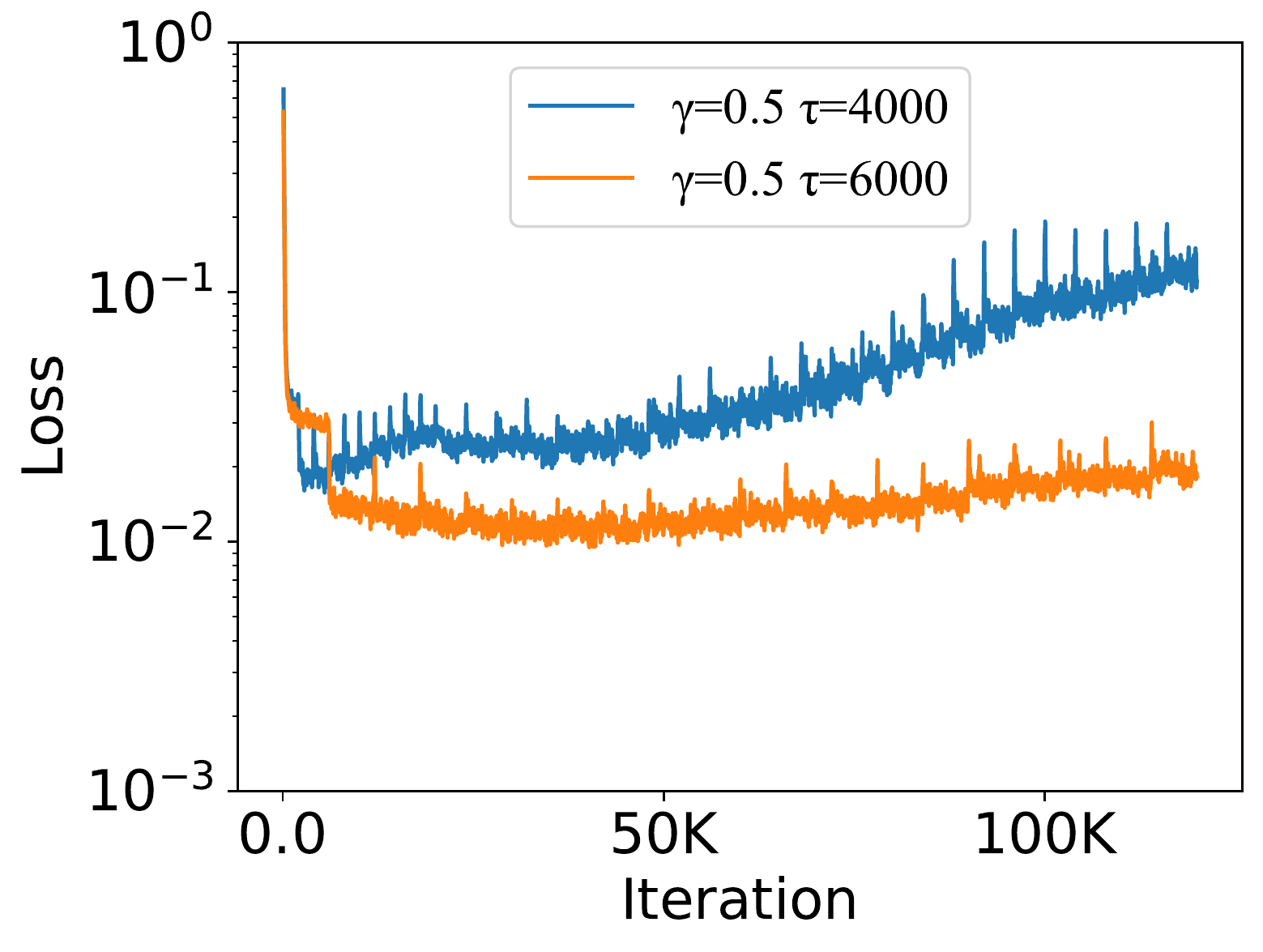}
  \label{fig:convergence_l}
\end{subfigure}
}
\caption{From left to right, convergence of the loss value during
  training for action spaces \highlevel{}, \midlevel{} and
  \lowlevel{}. Loss values are running means on logarithmic scale. As
  the size of the action space increases, the quality of the fit
  achieved by our model decreases. For action space \lowlevel{}, the loss value diverges even despite increasing parameter $\tau$.}
\label{fig:convergence}
\end{figure*}

To measure the quality of the fit achieved by our agents we record the
mean value of the loss function for sampled batches of experiences
throughout training. Figure~\ref{fig:convergence} shows how the loss
converges in all three action spaces. We achieve the best fit in the
action space~\highlevel{} with the relatively high value of
$\gamma=0.9$ which allows the network to account for long-term rewards
when predicting the values of different actions. We use $\gamma=0.5$
and increase the value of $\tau$ for larger action spaces to stabilize
the training. While the loss converges in action space~\midlevel{}, it diverges in action space~\lowlevel{} in spite of larger values of $\tau$. The disadvantage of increasing $\tau$ is that the training time also increases proportionally. As can be observed from Figure~\ref{fig:convergence_l}, using larger values of $\tau$ in action space~\lowlevel{} stabilizes training. However, it also prohibitively increases training time and therefore we refrain from further experiments with even bigger values of $\tau$.

\subsection{Metrics}
\label{sub:metrics}
In order to evaluate the optimization potential of an agent we compare
its performance with that of LLVM's built-in O3 optimization
sequence. To do that, we first calculate the speedup achieved by an
agent on every source code in the dataset. Then we aggregate these
values across training and validation sets by computing geometric
means of speedup for source codes in the respective sets. We do similar calculations for LLVM's O3 sequence and compare the computed metrics to evaluate the performance of an agent.

For an agent to learn the values of taking different actions, these
actions have to be explored first. As the agent continuously explores its environment, it accumulates new experiences which potentially yield higher speedups. In other words, highest observed speedups on source codes in the dataset continue to grow over time. These values put an upper bound on the agent's performance and enable us to tell how close it is to the best possible one. Therefore, during evaluation we also record the highest observed speedup for every source code in the dataset. Below we first present the results for aggregate metrics before showing the performance of our agents on individual source codes.

{ 
\renewcommand{\arraystretch}{1.15}
\setlength{\tabcolsep}{12.0pt}
\begin{center}
  \begin{table*}[h]
\centering
\caption{Top 5 best and worst performances on individual programs of an agent trained in action space~\highlevel{}.}
\label{tab:performance_by_source}
\footnotesize
\begin{tabular}[t]{lllrrc}
\toprule
           &           &                                 & \multicolumn{3}{c}{Speedup} \\
Dataset & Source Code & Action Sequence                  &      O3 &   Agent & Agent vs O3 \\
\midrule
\multirow{10}{*}{Training} & floyd-warshall.c & 4$\rightarrow$7$\rightarrow$7$\rightarrow$6$\rightarrow$6$\rightarrow$6$\rightarrow$6$\rightarrow$6$\rightarrow$6$\rightarrow$6$\rightarrow$4$\rightarrow$6$\rightarrow$6$\rightarrow$6$\rightarrow$6$\rightarrow$6 &   4.55x &   3.19x &       0.70x \\
           & reg\_detect.c & 5$\rightarrow$5$\rightarrow$0$\rightarrow$5$\rightarrow$1$\rightarrow$0$\rightarrow$0$\rightarrow$0$\rightarrow$0$\rightarrow$0$\rightarrow$0$\rightarrow$0$\rightarrow$0$\rightarrow$0$\rightarrow$0$\rightarrow$0 &   4.81x &   3.61x &       0.75x \\
           & flops-5.c & 4$\rightarrow$7$\rightarrow$7$\rightarrow$4$\rightarrow$4$\rightarrow$4$\rightarrow$7$\rightarrow$7$\rightarrow$7$\rightarrow$4$\rightarrow$4$\rightarrow$7$\rightarrow$4$\rightarrow$4$\rightarrow$4$\rightarrow$4 &   1.55x &   1.25x &       0.81x \\
           & fdtd-apml.c & 4$\rightarrow$4$\rightarrow$1$\rightarrow$4$\rightarrow$4$\rightarrow$1$\rightarrow$7$\rightarrow$7$\rightarrow$1$\rightarrow$7$\rightarrow$1$\rightarrow$1$\rightarrow$1$\rightarrow$1$\rightarrow$1$\rightarrow$1 &   1.13x &   0.95x &       0.84x \\
           & FloatMM.c & 2$\rightarrow$4$\rightarrow$0$\rightarrow$2$\rightarrow$2$\rightarrow$2$\rightarrow$6$\rightarrow$6$\rightarrow$6$\rightarrow$5$\rightarrow$6$\rightarrow$6$\rightarrow$6$\rightarrow$6$\rightarrow$1$\rightarrow$1 &   4.56x &   3.88x &       0.85x \\
           & ary3.c & 1$\rightarrow$6$\rightarrow$1$\rightarrow$1$\rightarrow$1$\rightarrow$1$\rightarrow$1$\rightarrow$1$\rightarrow$1$\rightarrow$1$\rightarrow$1$\rightarrow$1$\rightarrow$1$\rightarrow$1$\rightarrow$1$\rightarrow$1 &   5.45x &   6.85x &       1.26x \\
           & himenobmtxpa.c & 1$\rightarrow$4$\rightarrow$4$\rightarrow$0$\rightarrow$5$\rightarrow$0$\rightarrow$0$\rightarrow$3$\rightarrow$0$\rightarrow$3$\rightarrow$3$\rightarrow$0$\rightarrow$0$\rightarrow$0$\rightarrow$0$\rightarrow$0 &   1.38x &   1.82x &       1.32x \\
           & perlin.c & 0$\rightarrow$4$\rightarrow$6$\rightarrow$2$\rightarrow$2$\rightarrow$4$\rightarrow$2$\rightarrow$2$\rightarrow$2$\rightarrow$4$\rightarrow$2$\rightarrow$2$\rightarrow$2$\rightarrow$2$\rightarrow$4$\rightarrow$2 &   1.85x &   2.55x &       1.38x \\
           & puzzle.c & 1$\rightarrow$6$\rightarrow$2$\rightarrow$7$\rightarrow$3$\rightarrow$7$\rightarrow$3$\rightarrow$7$\rightarrow$2$\rightarrow$7$\rightarrow$3$\rightarrow$2$\rightarrow$7$\rightarrow$2$\rightarrow$2$\rightarrow$4 &   6.80x &  12.82x &       1.89x \\
           & Bubblesort.c & 0$\rightarrow$4$\rightarrow$7$\rightarrow$5$\rightarrow$5$\rightarrow$3$\rightarrow$3$\rightarrow$3$\rightarrow$0$\rightarrow$6$\rightarrow$6$\rightarrow$3$\rightarrow$3$\rightarrow$3$\rightarrow$6$\rightarrow$6 &   1.35x &   2.74x &       2.03x \\
\cline{1-6}
\multirow{10}{*}{Validation} & recursive.c & 1$\rightarrow$6$\rightarrow$6$\rightarrow$6$\rightarrow$6$\rightarrow$6$\rightarrow$6$\rightarrow$6$\rightarrow$6$\rightarrow$6$\rightarrow$6$\rightarrow$5$\rightarrow$6$\rightarrow$6$\rightarrow$0$\rightarrow$0 &   2.96x &   1.47x &       0.50x \\
           & dt.c & 7$\rightarrow$6$\rightarrow$2$\rightarrow$0$\rightarrow$0$\rightarrow$0$\rightarrow$0$\rightarrow$1$\rightarrow$1$\rightarrow$0$\rightarrow$0$\rightarrow$0$\rightarrow$0$\rightarrow$0$\rightarrow$0$\rightarrow$0 &   2.00x &   1.01x &       0.51x \\
           & fib2.c & 0$\rightarrow$2$\rightarrow$6$\rightarrow$2$\rightarrow$2$\rightarrow$2$\rightarrow$2$\rightarrow$2$\rightarrow$2$\rightarrow$2$\rightarrow$2$\rightarrow$2$\rightarrow$2$\rightarrow$2$\rightarrow$2$\rightarrow$2 &   1.65x &   0.96x &       0.58x \\
           & oourafft.c & 3$\rightarrow$4$\rightarrow$6$\rightarrow$3$\rightarrow$3$\rightarrow$3$\rightarrow$3$\rightarrow$3$\rightarrow$3$\rightarrow$3$\rightarrow$3$\rightarrow$3$\rightarrow$3$\rightarrow$3$\rightarrow$3$\rightarrow$3 &   2.99x &   1.79x &       0.60x \\
           & ackermann.c & 6$\rightarrow$7$\rightarrow$3$\rightarrow$6$\rightarrow$6$\rightarrow$7$\rightarrow$7$\rightarrow$7$\rightarrow$3$\rightarrow$6$\rightarrow$7$\rightarrow$6$\rightarrow$1$\rightarrow$2$\rightarrow$2$\rightarrow$2 &   6.95x &   5.87x &       0.84x \\
           & durbin.c & 5$\rightarrow$0$\rightarrow$4$\rightarrow$0$\rightarrow$0$\rightarrow$0$\rightarrow$0$\rightarrow$0$\rightarrow$0$\rightarrow$0$\rightarrow$0$\rightarrow$0$\rightarrow$0$\rightarrow$0$\rightarrow$0$\rightarrow$0 &   1.80x &   1.81x &       1.01x \\
           & pi.c & 1$\rightarrow$6$\rightarrow$5$\rightarrow$1$\rightarrow$5$\rightarrow$1$\rightarrow$5$\rightarrow$6$\rightarrow$5$\rightarrow$5$\rightarrow$5$\rightarrow$5$\rightarrow$6$\rightarrow$5$\rightarrow$6$\rightarrow$5 &   1.28x &   1.38x &       1.08x \\
           & flops-7.c & 0$\rightarrow$1$\rightarrow$2$\rightarrow$2$\rightarrow$2$\rightarrow$0$\rightarrow$1$\rightarrow$1$\rightarrow$1$\rightarrow$1$\rightarrow$1$\rightarrow$1$\rightarrow$1$\rightarrow$1$\rightarrow$1$\rightarrow$1 &   1.00x &   1.23x &       1.23x \\
           & jacobi-1d-imper.c & 1$\rightarrow$0$\rightarrow$7$\rightarrow$6$\rightarrow$5$\rightarrow$3$\rightarrow$3$\rightarrow$3$\rightarrow$3$\rightarrow$2$\rightarrow$3$\rightarrow$5$\rightarrow$3$\rightarrow$0$\rightarrow$0$\rightarrow$0 &   2.47x &   3.03x &       1.23x \\
           & dynprog.c & 4$\rightarrow$5$\rightarrow$4$\rightarrow$5$\rightarrow$0$\rightarrow$5$\rightarrow$5$\rightarrow$7$\rightarrow$3$\rightarrow$5$\rightarrow$1$\rightarrow$3$\rightarrow$3$\rightarrow$3$\rightarrow$3$\rightarrow$3 &   2.91x &   3.85x &       1.32x \\

\bottomrule
\end{tabular}
\end{table*}

\end{center}

}

\subsection{Aggregate Results}
\label{sub:aggregate_results}
\begin{figure*}%
\centering
\scalebox{.8}{

\begin{subfigure}{.32\textwidth}
  \centering
  \caption{Space \highlevel{}, training, $\delta$=4K}
  \includegraphics[width=1\linewidth]{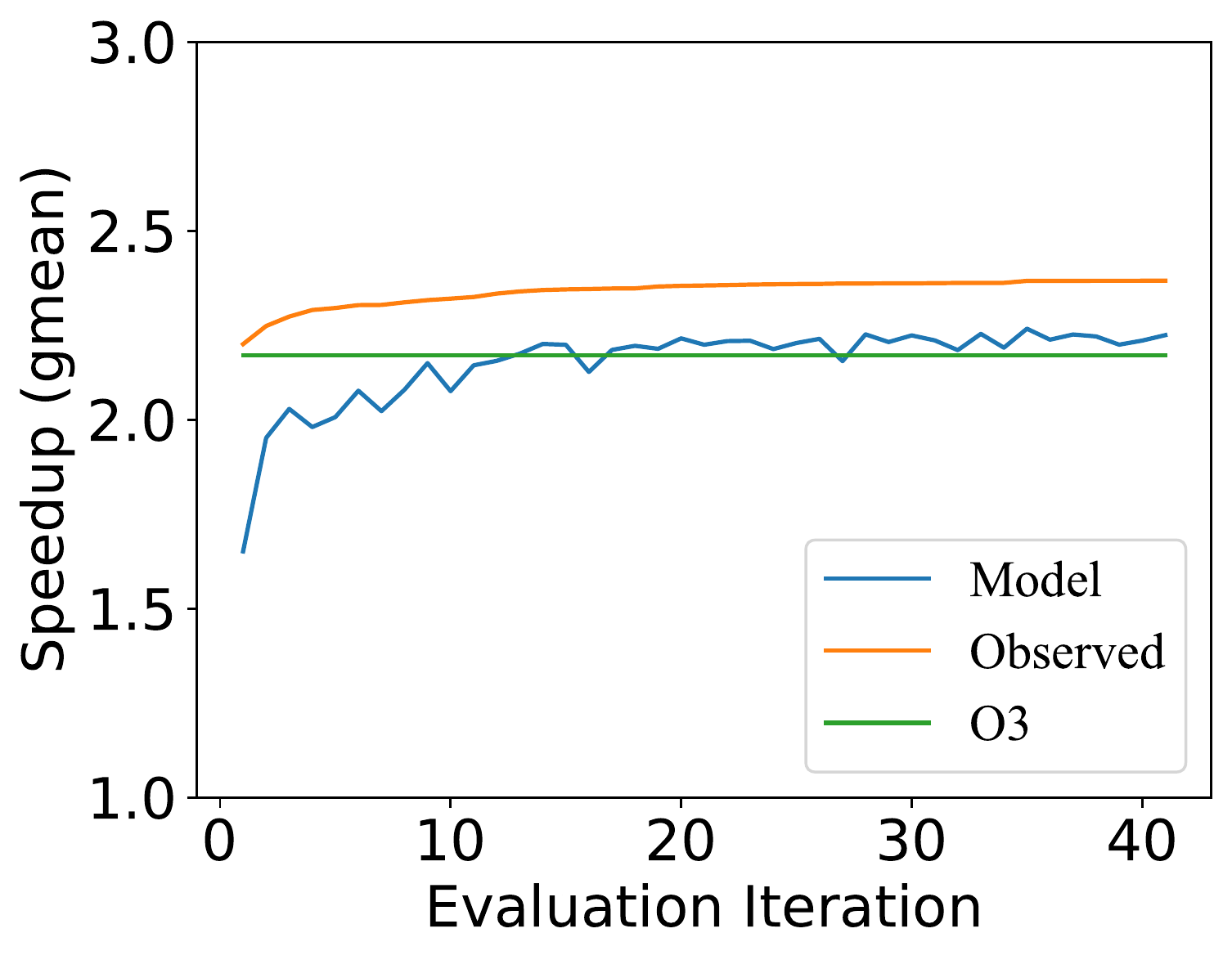}
  \label{fig:speedup_h_training}
\end{subfigure}\hspace{0.05\textwidth}
\begin{subfigure}{.32\textwidth}
  \centering
  \caption{Space \midlevel{}, training, $\delta$=10K}
  \includegraphics[width=1\linewidth]{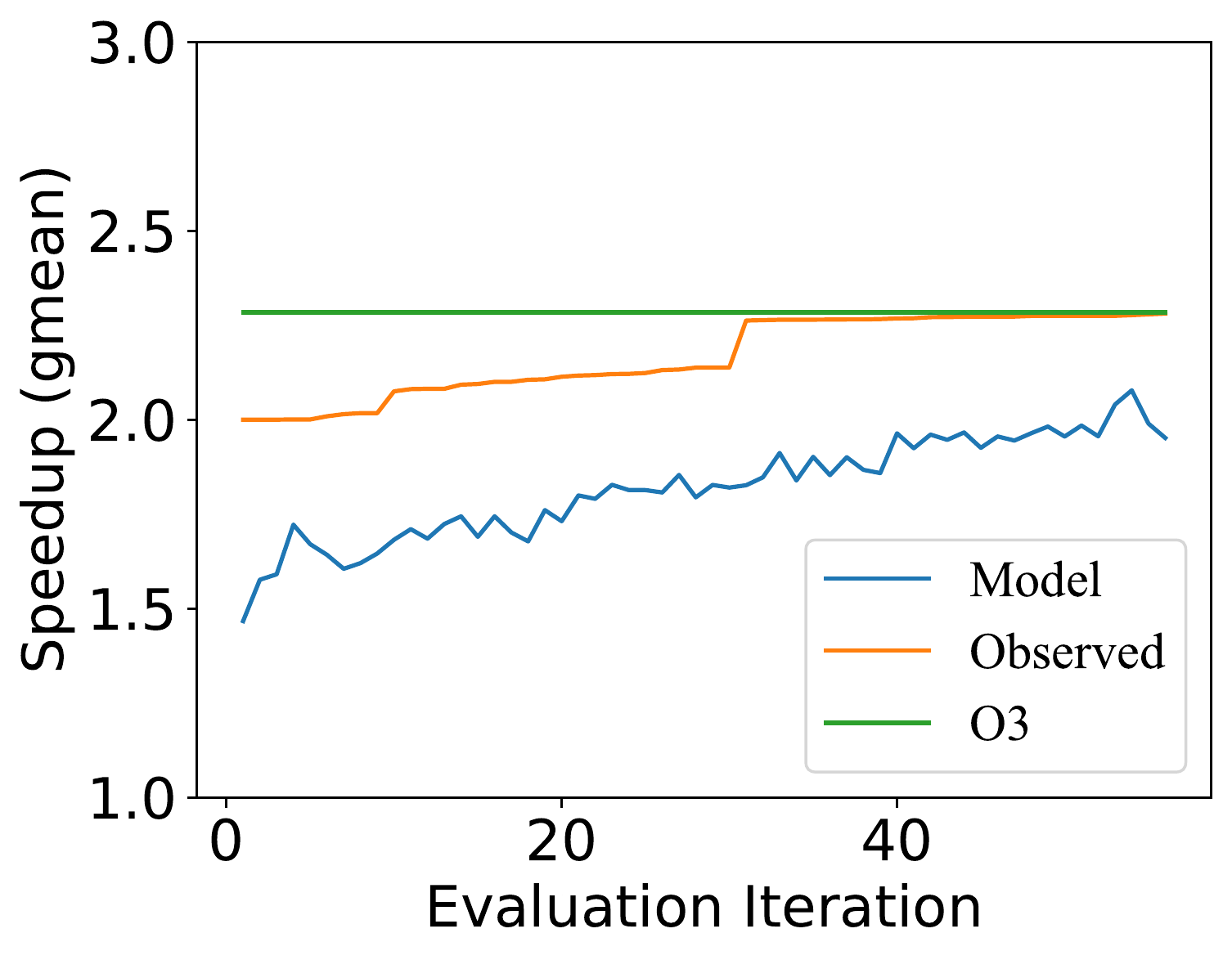}
  \label{fig:speedup_m_training}
\end{subfigure}\hspace{0.05\textwidth}
\begin{subfigure}{.32\textwidth}
  \centering
  \caption{Space \lowlevel{}, training, $\delta$=20K}
  \includegraphics[width=1\linewidth]{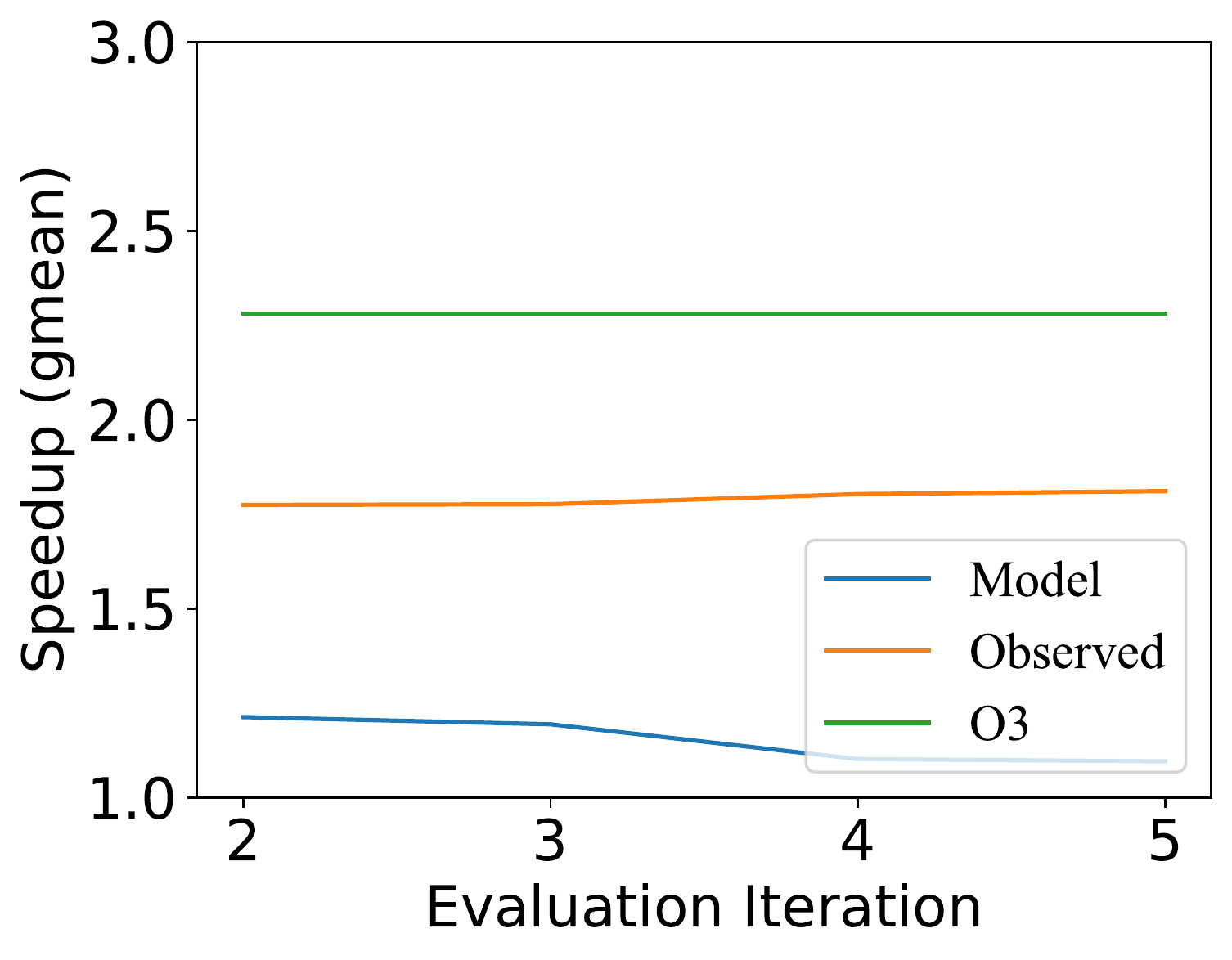}
  \label{fig:speedup_l_training}
\end{subfigure}
}

\scalebox{.8}{
\begin{subfigure}{.32\textwidth}
  \centering
  \caption{Space \highlevel{}, validation, $\delta$=4K}
  \includegraphics[width=1\linewidth]{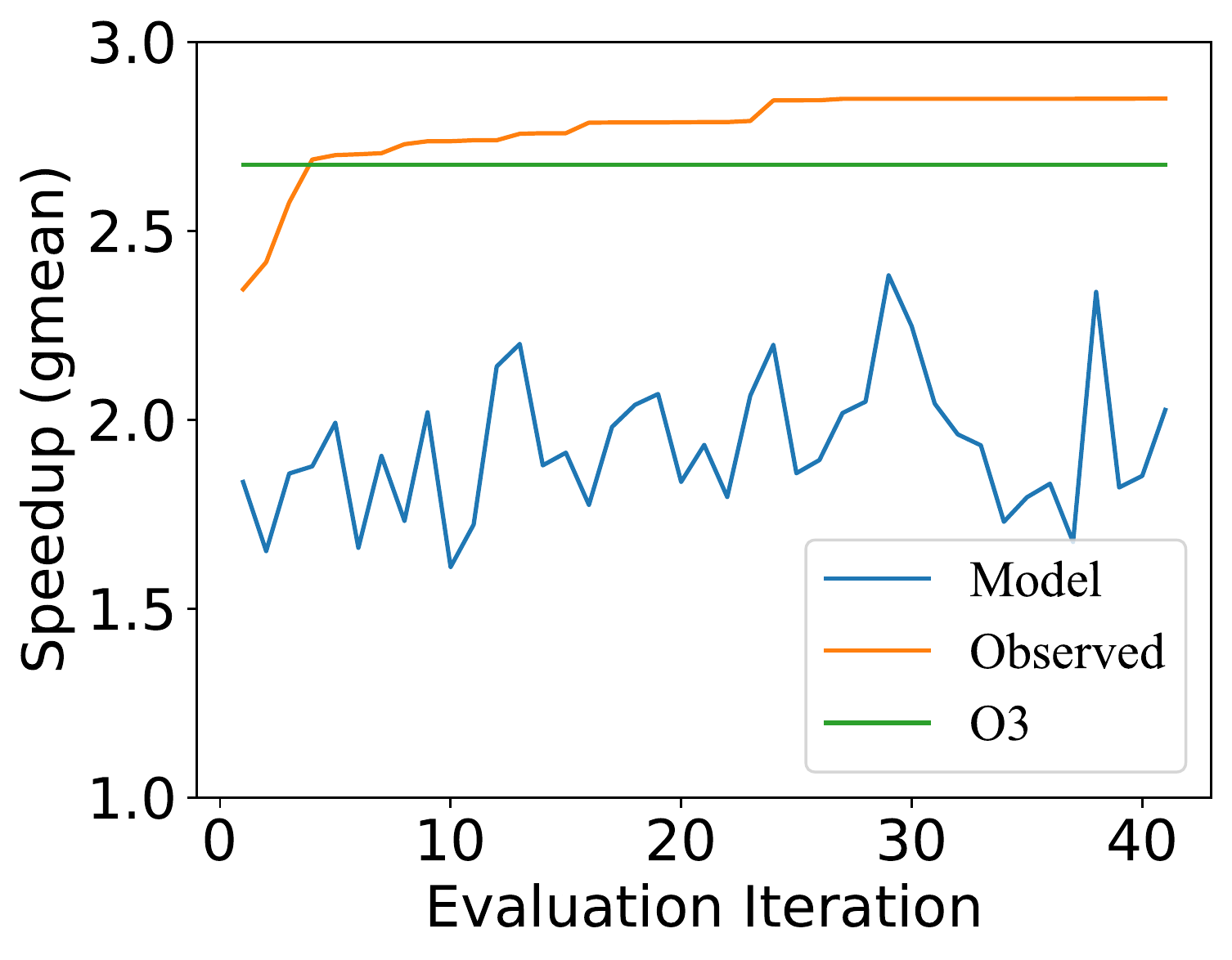}
  \label{fig:speedup_h_validation}
\end{subfigure}\hspace{0.05\textwidth}
\begin{subfigure}{.32\textwidth}
  \centering
  \caption{Space \midlevel{}, validation, $\delta$=10K}
  \includegraphics[width=1\linewidth]{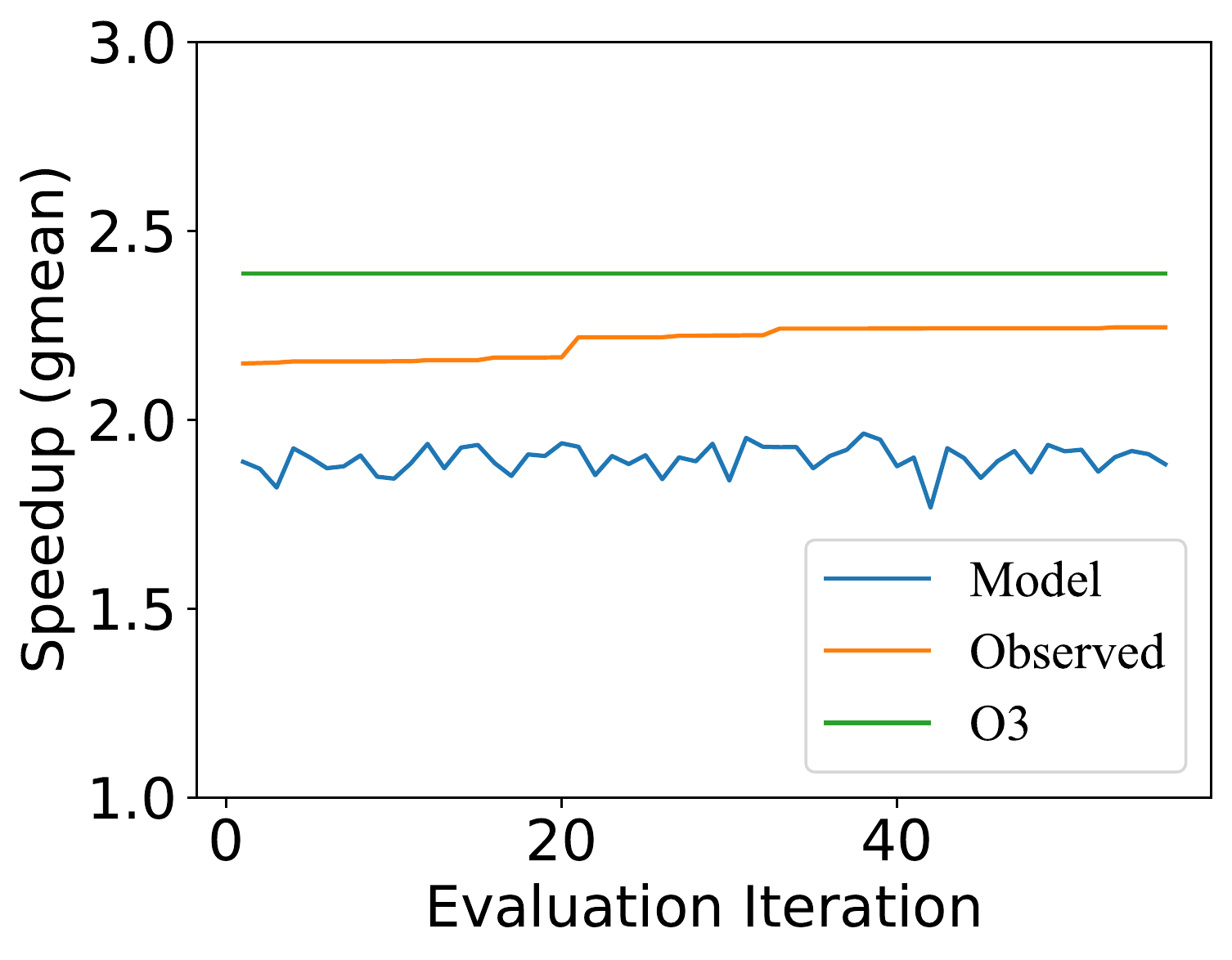}
  \label{fig:speedup_m_validation}
\end{subfigure}\hspace{0.05\textwidth}
\begin{subfigure}{.32\textwidth}
  \centering
  \caption{Space \lowlevel{}, training, $\delta$=20K}
  \includegraphics[width=1\linewidth]{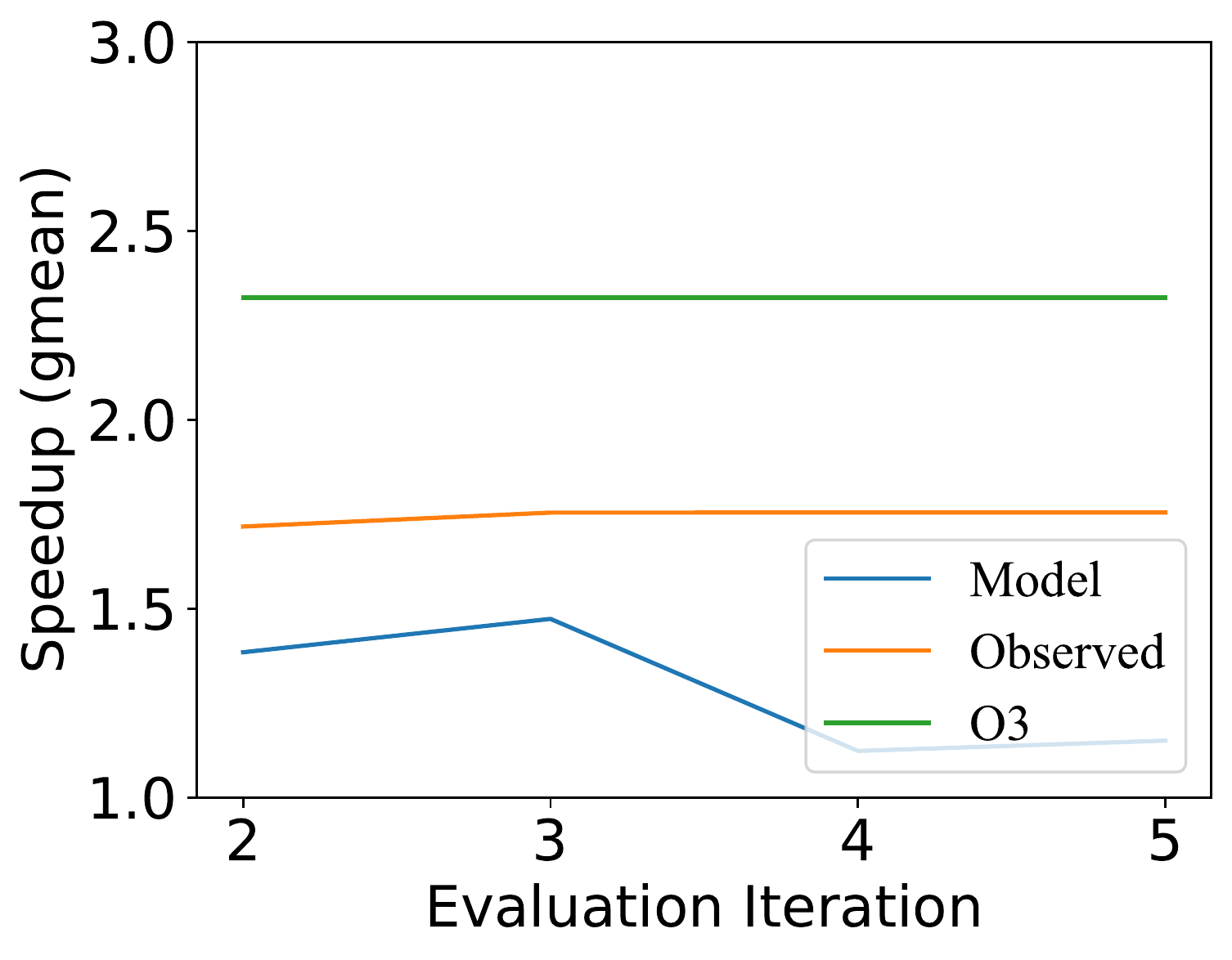}
  \label{fig:speedup_l_validation}
\end{subfigure}
}
\caption{Aggregate speedups in all three action spaces. The three curves in every plot show the average performance of a model, the average of the best observed performance on every program in the specified set and the average performance of LLVM's O3 sequence.}
\label{fig:speedups_aggregate}
\end{figure*}

As can be observed in Figure~\ref{fig:speedup_h_training} the agent
learns to outperform the O3 strategy on the training set in action
space~\highlevel{}, achieving an average speedup of 2.24x over the
unoptimized version, while the O3 sequence achieves an average
speedup of around 2.17x. The agent's performance is nearly 95\% of the
observed best-possible performance, which confirms that the model
achieves a good fit on the training
data. Figure~\ref{fig:speedup_h_validation} shows that, while the
validation set performance also increases over time, it only approaches
the performance of the O3 strategy, achieving an average speedup of 2.38x
over the unoptimized version versus the 2.67x average speedup achieved by O3. The growing best-observed performance on the validation set
shows that by behaving greedily the agent independently discovers
states corresponding to IRs with lower runtime than those produced by
O3. As we will see later, while the agent seldom
significantly outperforms the O3 strategy, it fails to be equally
robust across all the source codes. We attribute this mainly to a lack
of diversity in the distribution of source codes in our training set
and believe that having a larger more diverse training set would
likely solve the issue. Nonetheless, that we were able to discover the
IRs with lower runtime by re-arranging sub-sequences of passes
comprising LLVM's O3 routine shows that it is far from
optimal.

At every step in action space~\midlevel{}, our agent has to choose one
of 42 actions, each corresponding to a particular LLVM pass. Since
this space is much larger than space~\highlevel{} it takes much longer
for the agent to discover advantageous
states. Figure~\ref{fig:speedup_m_training} shows that after
more than forty evaluation steps, which includes nearly six days of exploration within that period,
the agent is able to observe experiences yielding the same average
speedup over the baseline as the O3 sequence. During this time the
agent continuously improves its performance on the training set and
given enough time is likely to achieve and surpass the performance of
the O3 strategy. However, its performance on the validation set does
not seem to improve as seen in
Figure~\ref{fig:speedup_m_validation}. Therefore, in view of the limited access to compute resources, we terminate the experiment in action space~\midlevel{} after 56 evaluations. We believe that given a larger number of actions in space~\midlevel{} when compared to~\highlevel{} it is easier for the agent to memorize action sequences yielding high speedups on specific source codes in the training set. Similar to action space~\highlevel{}, increasing the size and diversity of the training set is likely to force the agent to generalize and achieve better performance on the validation set.

Given that in our experiments in the action space~\lowlevel{} the loss
function diverges, we do not see any meaningful improvement in agent's
performance during the evaluation, as shown in Figures~\ref{fig:speedup_l_training} and~\ref{fig:speedup_l_validation}.

\subsection{Performance on Individual Programs}
\label{sub:individual_performance}

To examine the behavior of an agent trained in action
space~\highlevel{} on individual source codes, we record the sequence
of actions chosen by the model for every program. This allows us to
verify that the model does indeed produce a different optimization
strategy for different programs. Furthermore, we calculate the speedup
achieved by our agent and LLVM's O3 strategy over the unoptimized base
version of the IR of every source code in our
dataset. Table~\ref{tab:performance_by_source} presents top five best
and worst performance results in both training and validation sets.

By observing the results we can conclude that the agent does indeed
produce specialized optimization strategy for every source
code. Interestingly, the agent utilizes the balance of $\mu=16$
available actions to the fullest in almost all cases, except some that are not shown in Table~\ref{tab:performance_by_source}. This means that
in most cases the agent predicts at least one action to yield positive reward.
Although the IR itself does not necessarily change as a
result of every action, the history of actions is always updated to
store the latest action. Since the state consists of both the IR and
the history of actions, it changes after every action of the agent. Therefore we stop sampling the agent only when all of the actions are predicted to lead to slowdown, i.e., have value 0 or less, or after the maximum number of actions $\mu$ is taken.

\section{Related Work}
\label{sec:related}

Compiler optimization problem has been in focus of research community for several decades, with earliest works dating back to late 1970s~\cite{leverett1979overview}. Its subproblems of various complexity, ranging from the simplest, parameter value selection, to the most complex, phase-ordering, were tackled via different classes of methods~\cite{ashouri2018survey}. Among these methods are iterative search techniques ~\cite{bodin1998iterative}, genetic algorithms~\cite{cooper1999optimizing, cooper2002adaptive, kulkarni2004fast}, and machine learning methods~\cite{fursin2011milepost, kulkarni2012mitigating, Ashouri:2017} with deep learning methods gaining popularity in recent years~\cite{Cummins2017a, cummins2020programl}.

In order to leverage the advantages of (deep) machine learning methods when it comes to compiler optimization, several challenges need to be addressed: (i) correctly defining the learning problem, (ii) choosing or building the right set of features to represent the program, (iii) generating the dataset for training, and (iv) selecting the right neural network architecture which is both expressive enough to learn the task and allows efficient training. The learning problem is defined as either an unsupervised learning problem, often used to learn features~\cite{Cummins2017a, NIPS2018, brauckmann2020compiler, cummins2020programl}, a supervised learning problem~\cite{Cummins2017a, Ashouri:2017}, or a reinformcement learning problem~\cite{kulkarni2012mitigating}. A set of features includes statically-available ones, such as code token sequences~\cite{Allamanis2013, Cummins2017, Cummins2017a},  abstract syntax trees (AST) and AST paths~\cite{dam2018deep, Alon2018, Alon2018a}, IRs and learned representations built on top of IRs~\cite{aggarwal2019ir2vec, NIPS2018, brauckmann2020compiler, cummins2020programl, Park:2012, Allamanis2017b}. An additional set of features includes the problem size~\cite{Cummins2017a} and dynamic performance counters~\cite{cavazos2007rapidly}. Training data is often generated manually for supervised learning methods~\cite{Cummins2017a, Ashouri:2017}, while reinforcement learning methods use initial training set to generate data via exploration~\cite{kulkarni2012mitigating}. Unsupervised learning methods can take the advantage of the large code corpora available online~\cite{NIPS2018, cummins2020programl}. There also exist methods for automatic generation of training data using deep neural networks~\cite{Cummins2017}.

Our work is most similar to the approach by Kulkarni et al.~\cite{kulkarni2012mitigating}, who also use reinforcement learning and train a neural network to tackle the phase-ordering problem. However, important differences from the above work are the following: (i) our approach does not depend on dynamic features and therefore does not require a program to be run to make a prediction, (ii) the search space of possible optimizations considered in our work is much larger, (iii) our approach depends on the IR of the program and is therefore agnostic to the front-end language a program is written in, and (iv) instead of NEAT, we use gradient-based optimization to train our neural network.

Ashouri et al.~\cite{Ashouri:2017} developed the MiCOMP framework to tackle the phase-ordering problem by first clustering LLVM passes composing the O3 sequence of the LLVM optimizer and then using a supervised learning approach to devise an iterative compilation strategy which outperforms the O3 sequence within several trials. Similar to Kulkarni et al.~\cite{kulkarni2012mitigating}, they use dynamic features and consider a smaller search space of size $5^6$ compared to $8^{16}$, which is the size of \highlevel{}, the smallest action space considered in our work.

%
%
%
%
%

\section{Conclusion}
\label{sec:conclusion}

We formulated compiler phase-ordering as a deep reinforcement learning problem and developed the~\Name{} framework, which allows for efficient training of optimizing agents. Our approach is fully automatic and relies only on the initial supply of a dataset of programs. We were able to train the agents which surpass the performance of LLVM's hard-coded O3 optimization sequence on the observed set of source codes and achieve competitive performance on the validation set, gaining up to 1.32x speedup over the O3 sequence with previously unseen programs. We believe these results exhibit the big potential of deep reinforcement learning in tackling phase-ordering problem of compilers.

Our approach has several shortcomings, which we plan to address in the future.
Firstly, increasing the size of the dataset to include a more diverse set of source programs might be enough to achieve superior performance compared with the hard-coded optimization strategy. Secondly, using higher-quality embeddings for the IR and the appropriate neural architecture can result in more efficient and robust optimizing agents. 
Next, current design requires that the programs are compiled and benchmarked on every new target system, which requires substantial computational resources. While calculation of rewards by running the benchmarks on the end systems is at the center of our approach, we believe the data efficiency of the learning procedure could be improved by including a self-supervised learning step by the agent. This would potentially result in a more efficient exploration strategy, and reduce the computational burden by allowing faster convergence of an agent.
Finally, optimizing the agents' training procedure could allow for similar results to be achieved in higher dimensional action spaces.

\section*{Acknowledgments}

This work is supported by the Graduate School CE within the Centre for Computational Engineering at Technische Universität Darmstadt and by the Hessian LOEWE
initiative within the Software-Factory 4.0 project. The calculations
for this research were conducted on the Lichtenberg Cluster of TU
Darmstadt.

\bibliographystyle{IEEEtran}
\bibliography{IEEEabrv,bibfile}

\end{document}